\pdfoutput=1

\documentclass[11pt]{article}

\usepackage[]{acl}
\usepackage{graphicx}
\usepackage{amsfonts}
\usepackage{amsmath}
\usepackage{booktabs}
\usepackage{times}
\usepackage{latexsym}

\usepackage[T1]{fontenc}

\usepackage[utf8]{inputenc}

\usepackage{microtype}

\title{Poirot at SemEval-2022 Task 5 : Misogynistic Meme Detection using Early Fusion Model with Graph Network}

\author{Harshvardhan Srivastava \\
  Oracle India \\
  Indian Institute of Technology, Kharagpur\\
  \texttt{harshvardhan.srivastava@oracle.com}
}

\begin{document}
\maketitle
\begin{abstract}
In recent years , there has been an upsurge in a new form of entertainment medium called memes. These memes although seemingly innocuous have transcended onto the boundary of online harassment against women and created an unwanted bias against them . To help alleviate this problem , we propose an early fusion model for prediction and identification of misogynistic memes and its type in this paper for which we participated in SemEval-2022 Task 5 . The model receives as input meme image with its text transcription with a target vector.
 Given that a key challenge with this task is the combination of different modalities to predict misogyny, our model relies on pretrained contextual representations from different state-of-the-art transformer-based language models and pretrained image pretrained models to get an effective image representation .Our model achieved competitive results on both SubTask-A and SubTask-B with the other competition teams and significantly outperforms the baselines. 
\end{abstract}

\section{Introduction}

Meme culture in today's virtual climate gives us a variety of insight into the pop culture, general ideology and lingustic conversational manner of the generation. To understand the internet culture , it becomes essential to study memes \cite{memedefinition} and the impact it has on the internet people. Some of the most popular communication tools in social media platforms are memes. Memes are essentially images characterized by the content of a picture overlaid with  text that was introduced by people with the main purpose of being interesting and ironic. Women have a strong presence online, especially on image-based social media like Twitter , SnapChat, Instagram. 78\% of women use social media several times a day, compared to 65\% of men. While new opportunities are being opened up for women online, systematic inequality and discrimination are being replicated offline from these online spaces in the form of offensive content for women. Most of them were created with the intention of making funny jokes, but soon people began to use them as a form of hatred for women, leading to sexist and offensive messages in the online environment, and as a consequence,  the sexual stereotyping and gender inequality of the offline world where sexuality stereotypes and gender inequality have been strengthened. This insensitive and obscene type of meme has a profound effect on a person's mental health and can exhibit harmful effects on cognitive and emotional processes leading to mental illnesses as shown in \citet{sexistmeme}.

In this work, we present team Poirot's solution to SemEval - 2022 Task 5 competition as described in detail in \citet{task5}. We focused our efforts on our primary approach of building a Multi-Modal-Multi-Task module that uses features from both images and text.  Furthermore, in this paper , we provide ablation studies on different modalities , relative importance of the different modalities and some training parameters, and show how by changing the module parameters , the predictions on the misogynistic identification of memes aggrevates or allays.

\section{Background}
\subsection{Task Description }
The organisers have provided us with data tasked with the identification of misogynous memes, taking advantage of both text and images available as source of information. The task is comprised around two main sub-tasks:
\begin{itemize}
\item Sub-Task A: a basic task about misogynous meme identification, where a meme should be categorized either as misogynous or not misogynous;
\item Sub-Task B: an advanced task, where the type of misogyny should be recognized among potential overlapping categories such as stereotype, shaming, objectification and violence.
\end{itemize}
The sub-tasks are arranged in increasing range of difficulty . The competition is challenging, as identifying the misogynous nature of a meme is more complex in a multi-modal setting than performing the same task only on textual data. For memes, comprised of image and text information, a multi-modal approach for understanding both visual and textual cues is needed. Also , in sub-task B , the nature of problem difficulty is increased as the type of misogyny has to be identified , which can belong to multiple categories due to the nature of the dataset .

\subsection{Dataset}

\begin{figure}
    \centering
    \includegraphics[width = 0.9\linewidth]{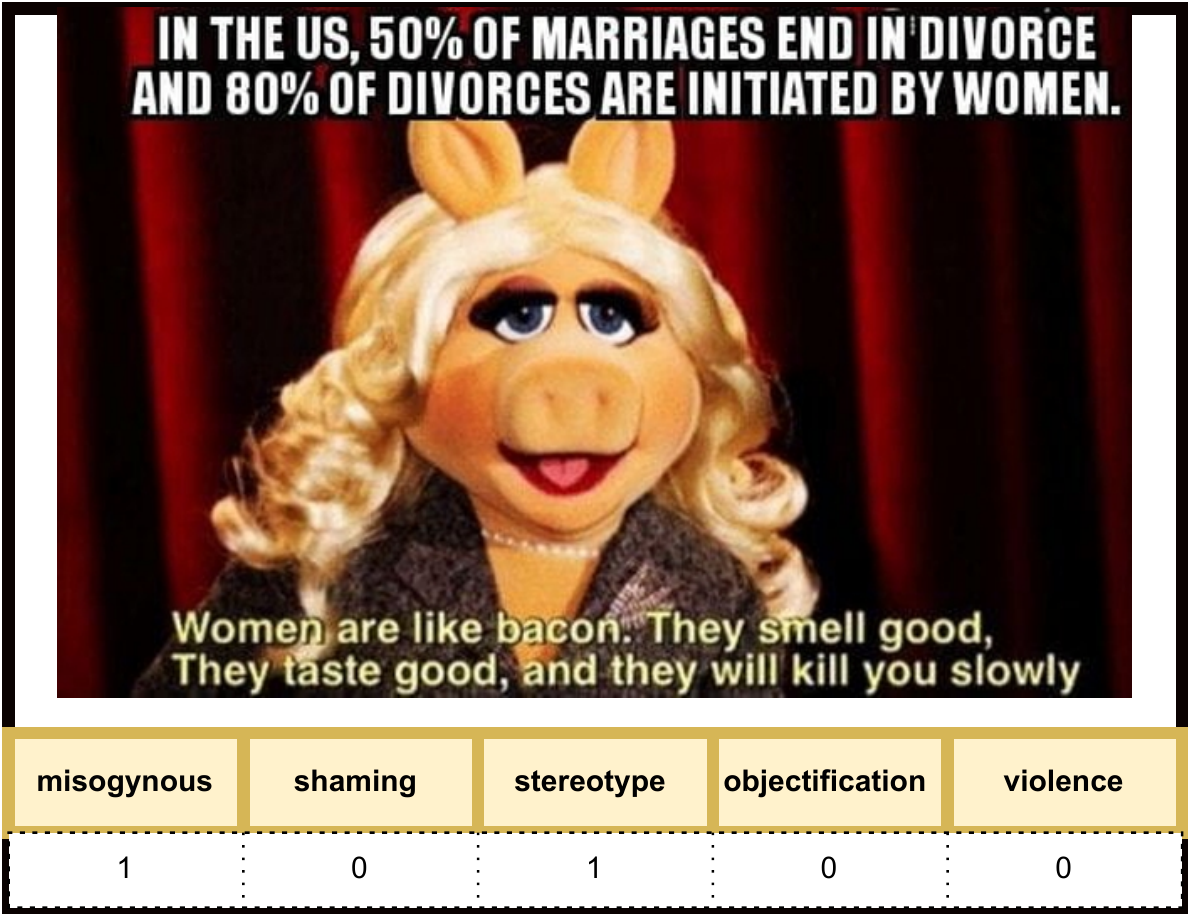}
    \caption{Misogyny in meme}
    \label{fig:datasample}
\end{figure}

The datasets for the competition provided by the task organisers are memes collected from the web and manually annotated via crowdsourcing platforms \cite{task5}. Each sample is supported by an image and the corresponding text transcription (if it exists) on the image . The statistical information about the datasets can be found in Table \ref{tab:datasetinfo}.

Additionally, we provide a quick look into the training dataset which has a significant data imbalance in number of samples belonging to each of the \textit{5} given labels. This imbalance affects the performance of the models on the test set specially in the case of {multilabel} prediction as not equal training instances are available for each of the classes. The information about the number of samples belong to each of the 5 classes for the training set is given in Table \ref{tab:datasetinfo}. This dataset imbalance is dealt with in section \ref{subsec:customloss}.

\subsection{Evaluation Criteria}
The teams’ performance is evaluated by the macro F1 score for task A. For tasks B ,
the weighted F1 score is computed for each subtask (misogynous, shaming, stereotype, objectification, violence), and the average F1 score of these subtasks is used to rank the systems.

\begin{table}[h]
\centering
\begin{tabular}{c}
\begin{tabular}{|l|c|}
\hline
\textbf{Set} & \textbf{Number of Samples}\\
\hline
\verb|Trial| & {100} \\
\verb|Train| & {10000} \\
\verb|Test| & {1000} \\ 
 \hline
\end{tabular} \vspace{4mm}
\cr
\begin{tabular}{|l|c|}
\hline
\textbf{Label} & \textbf{Positive Samples}\\
\hline
 {\textit{misogynous}}& {5000} \\
{\textit{shaming}} & {1274} \\
{\textit{stereotype}} & {2810} \\ 
{\textit{objectification}}& {2202} \\
{\textit{violence}}& {953} \\
 \hline
\end{tabular}
\end{tabular}
\caption{Dataset and Labels Information }
\label{tab:datasetinfo}
\end{table}

\begin{figure*}
\centering
\frame{\includegraphics[width = 0.95\textwidth]{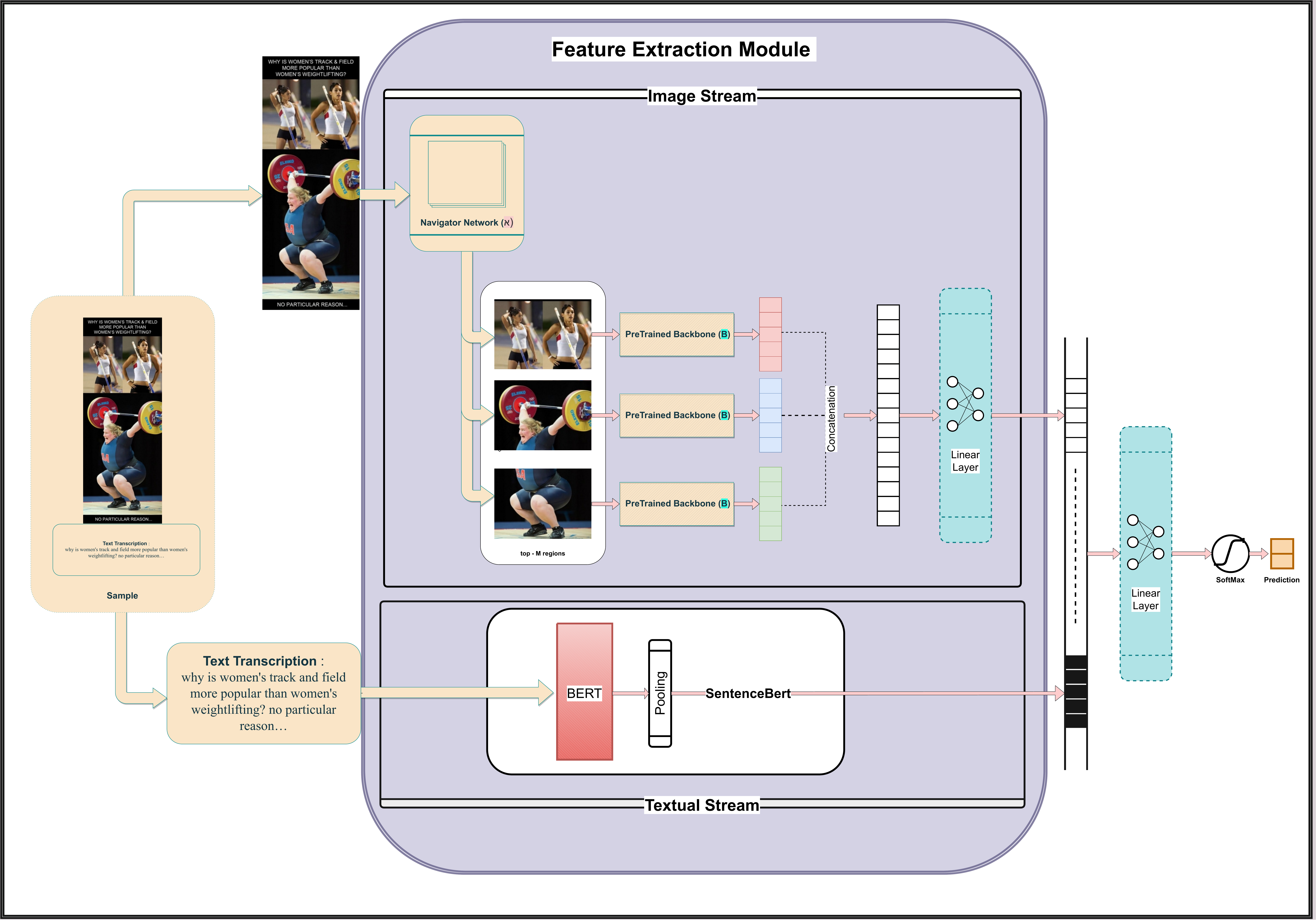}}
\caption{ Overview of the binary model used for misogyny detection. The two modalities are passed simultaneously through the Feature Extraction Module in two separate paths and trained together and finally fused together and passed through a linear layer to get binary logits .}
\label{fig:binary}
\end{figure*}

\section{System Overview}
The solution system comprises of 2 seperate systems for both the sub-tasks . The approach can be broadly divided into binary approach and multilabel approach .
\subsection{Binary Model}
\label{subsec:binary}
Broadly , the model consists of two modal information streams , text and image . The proposed approach leverages on multimodal information to provide the classification of a sample .We exploit text transcriptions written in natural language jointly with visual information coming from the meme image. In the initial stage the pipeline is divided into two streams running in parallel which on later stage is joint together. The outline of the proposed architecture is shown in Figure \ref{fig:binary}. The submodules are described below:

\begin{enumerate}
\item \textbf{Image Features Extraction} : This stream is also separated into two sub-modules :
\begin{itemize}
\item \emph{Pretrained Representation Module} :We can use any backbone CNN base models to learn the features of an image. For our experimentation , we use ResNet-101 and ResNet-50 \cite{resnet} as the backbone \textit{B} model . Thus if the input image $\textit{I} \in  \mathbb{R}^{d \times d}$ , where d = 224 , the output of the feature extractor would give us intermediate level features $\kappa \in  \mathbb{R}^{D}$ , where D = 2048 ,
\[\kappa = \textit{\textit{B}}(I) \]
\item \textit{Navigator Module} : We use Navigator Module from the NTS-net as described in \cite{ntsnet} model to decouple image into several parts. For an input image, the image is fed into Navigator network to compute the informativeness of all regions .It is fed to the navigator network , which extracts meaningful parts to separate \textit{top-M} regions . The feature extractor extracts its deep feature map for each of those parts . These features are then concatenated $C$ together: 
\[\kappa_{i} = \aleph (I)  , i \in [0,M-1]\]
\resizebox{\linewidth}{!}{$f_{v} = C (\kappa_{0}, \kappa_{1}, \kappa_{2},....,\kappa_{M-1}) \in \mathbb{R}^{M \times D} $}
\end{itemize}

\item \textbf{Text Features Extraction:} : The second modality being the textual Stream, uses the SentenceBERT model \cite{sentencebert}. SBert modifies the BERT network using a combination of siamese and triplet networks to derive semantically meaningful embedding of sentences. As a state of the art language model , BERT has greatly influenced results in the text classification task , we use SentenceBert $S$ model trained on Siamese BERT networks . Thus we convert the given text $T$ transcription into features vector $f_{t}$ . Formally : 
\\
\[f_{t} = S(T)   \in    \mathbb{R}^{E}\]
\\
where E = 768.
\end{enumerate}

The image extraction part and text extraction part is clubbed together to form the \textbf{Feature Extraction Module}.
\[F = [f_{v},f_{t}]\]
This module outputs a feature vector of size $N = E+D $ features. These features are then passed through a $f$(linear layer) to output logits which are then passed on to $\sigma$ function to generate predictions.

\[y_{pred} = \sigma(f(F)) \]

\subsection{Multi-Label Model}
The multi-label model , keeps the feature extracting pipeline of the network in binary model intact , while changing the final output method by using Graph Neural Networks . The model consists of two essential parts : (i) Feature Extraction Module and (ii) Graph classification module  .
\begin{enumerate}
\item \textbf{Feature Extraction Module} : Same as in binary model \ref{subsec:binary}
\item \textbf{Graph Classification Module} : Graph has an effective message passing system , which can be modelled to find the inter-dependency of the labels amongst each other , and hence , efficiently capture the semantic importance of a label $u_{1}$ depending on co-occurring label $u_{2}$. We represent each node of the graph input to be a label , having the node features as GloVe embedding having \textit{e} features . Formally , we use Graph Network to learn the multi-label classification model to learn label representation: 
\[L_{n+1} = \phi(L_{n},A)\] where $L_{n} \in \mathbb{R}^{u \times e}$ represents class label representation at \textit{nth} graph layer , $\phi$ represents the message passing network and $A \in \mathbb{R}^{u \times u}$ represents the adjacency matrix. Through stacking multiple Graph Network Layers, we model the complex inter-relationships among classes.

\begin{figure*}[ht]
\centering
\frame{\includegraphics[width = 0.9\textwidth]{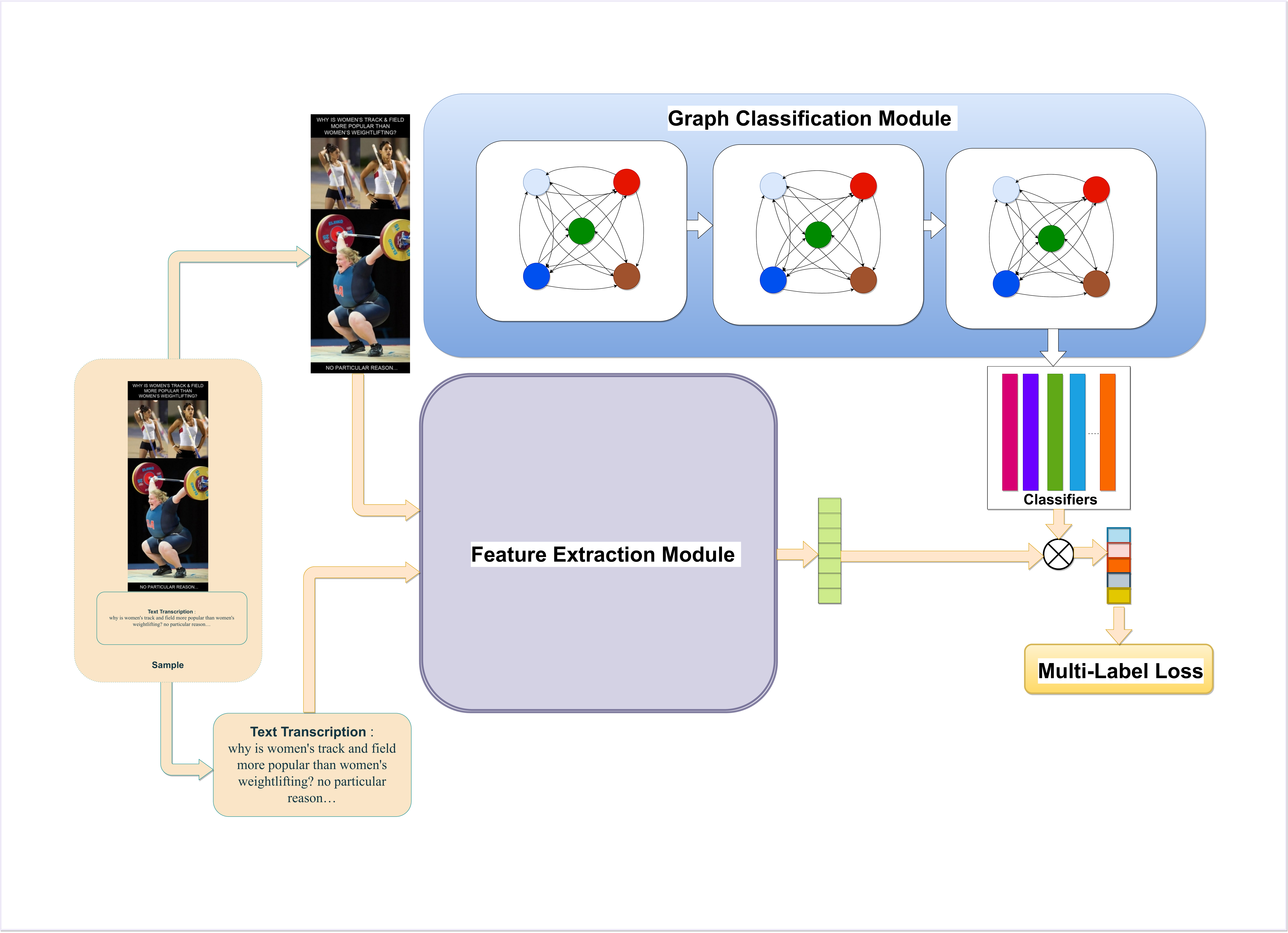}}
\caption{: The general architecture of our Multi-Modal-Multi-Label model. We use image features and text features from the \emph{Feature Extraction} (FE) which has a pretrained ResNet , pretrained sentence transformer SBert . The features are passed passed through a 5-layer classifier stack generated from the Graph Classification Module which takes input the label's semantic information to generate the output multi-label prediction.}
\end{figure*}

\textbf{Creation of Adjacency Matrix} : We calculate the label adjacency matrix $\mathbb{A}$ by mining label co-occurrence patterns in the training and trial dataset. Let the label matrix $L_{m} \in \mathbb{R}^{n_{s} \times u}$ , where $n_{s}$ are the number of training and trial samples . Then the co-occurrence matrix 
\[A_{coo} = L_{m}^{T} \times L_{m}  \in \mathbb{R}^{u \times u}\]To create the adjacency matrix from the co-occurrence matrix and to remove the self node loop from the graph , we create a vector $N_{u} \in \mathbb{R}^{u}$ , having  
\[N_{u}[i] =  A_{coo}[i][i]\]
Finally , the adjacency matrix $\mathbb{A}$ can be constructed as :
\\

\begin{equation*}
    A_{ij} = 
    \begin{cases} 0 & \text{ if $i = j$,}
    \\
    \frac{A_{coo}[i][j]}{N_{u}[i]} &\text{ otherwise}
    \end{cases}
    \end{equation*}
\end{enumerate}
\subsection{Multi-Label Classification Loss}
\label{subsec:customloss}
We notice the imbalance present in data for different classes which we can see in Table  \ref{tab:datasetinfo}, but the extent of imbalance is different for different labels. This knowledge can be passed to the neural network in terms of class weights in order to penalize adequately. Let $n_{s}$ be the number of samples in the dataset . We calculate the weighted importance of a class using the below equations:
\[
    W_{p}[i] = \frac{N_{p}[i]}{n_{s}} ,
    W_{n}[i] = \frac{N_{n}[i]}{n_{s}} 
\]
where $N_{p}[i]$ , is the number of positive samples for class \textit{i} and $N_{n}[i]$ , is the number of negative samples for class \textit{i}.
\[-dl = W_{p}*y*\log(p) + W_n*(1 - y)*\log(1 - p)\]
where y is the ground truth and p is the predicted output . The calculated weights are shown in table \ref{tab:weights}.

\begin{table}[]
    \centering
    \begin{tabular}{lcc}\toprule
         & \multicolumn{2}{c}{Weights}  \\\cmidrule{2-3}
         \textbf{Label} & \textbf{Positive ($W_{p}$)} & \textbf{Negative ($W_{n}$)} \\\midrule
         misogynous & 1.000 & 1.000 \\
         shaming & 3.924 &  0.573 \\
         stereotype & 1.779 & 0.695\\
         objectification & 2.270 & 0.641 \\
         violence & 5.246 & 0.552\\\bottomrule
    \end{tabular}
    \caption{Calculated weights for regularizing cross-entropy loss in the custom loss function}
    \label{tab:weights}
\end{table}

\section{Experimental Setup }
\subsection{Baselines}
We use the baseline provided by the task organisers \footnote{\href{https://github.com/MIND-Lab/MAMI}{https://github.com/MIND-Lab/MAMI}} which depend on the Sub-Task and use a different set of features for different tasks :
\begin{enumerate}
\item Sub-Task A : Misogynous Meme Identification
\begin{itemize}
\item {\textbf{\textit{Baseline-Text }}}: deep representation of text, i.e. a fine-tuned sentence embedding using the USE pre-trained model
\item {\textbf{\textit{Baseline-Image}}} : deep representation of image content, i.e. based on a fine-tuned image classification model grounded on VGG-16.
\item {\textbf{\textit{Baseline-IT}}} : concatenation of deep image and text representations, i.e. based a single layer neural network.
\end{itemize}
\item Sub-Task B : Type of Misogynous Meme Identification
\begin{itemize}
\item \textbf{\textit{Baseline-Flat }}: a multi-label model, based on the concatenation of deep image and text representations, for predicting simultaneously if a meme is misogynous and the corresponding type
\item \textbf{\textit{Baseline-Hierarchical}} : a hierarchical multi-label model, based on text representations, for predicting if a meme is misogynous or not and, if misogynous, the corresponding type.
\end{itemize}
\end{enumerate}

\subsection{Hyperparameters and Implementation Details}
Before passing the text transcription to the text stream , we apply some basic text pre-processing to all our sentences. First, we normalize all the sentences by converting all white-space characters to spaces. Also , in the image stream , before passing the image to the navigator network , we resize the image to size [224,224] for uniformity and perform random crops and flips before feeding to the network . When concatenating the image features with textual features , we use a parameter $\lambda$ to combine the two together.We adopt a 2-layer graph network for our best performing system .For node features , we use the 300-Dimensional GloVe embeddings trained on the Wikipedia Dataset .Table \ref{tab:hyperparameters} contains the list of general hyperparameters we used . We implement the network based on PyTorch.

\begin{table}[h]
\centering 
\begin{tabular}{c c} 
\hline\hline 
Parameter Name & Value \\ [0.5ex] 
\hline 
Optimizer & AdamW  \\ 
Pre-Trained BERT LR & 2e-4  \\
Navigator Network LR & 1e-3 \\
Graph Learning Rate & 1e-2  \\
Graph Layer 1 Dim & 512  \\ 
Graph Layer 2 Dim & 2048 \\ 
$\lambda$ (Concatenation Parameter) & 0.7 \\ [1ex]
\hline 
\end{tabular}
\caption{Major hyperparameters used} 
\label{tab:hyperparameters} 
\end{table}

\section{Results}
Table \ref{tab:binresults} and \ref{tab:multiresults} compares the macro and weighted $f_{1}$ scores of our best performing models on the binary classification task and the multi-label task respectively alongside the score achieved by the baseline models. We also present several ablations for the best performing models on $\lambda$ parameter and its effect on the final score achieved by the model. The $\lambda$ ablations can be found in Table \ref{tab:lambda}.

ResNet-101$_{imagenet}$ uses the backbone \emph{B} pre-trained on the ImageNet dataset , while ResNet-50$_{nsfw}$\footnote{\href{https://github.com/emiliantolo/pytorch_nsfw_model}{https://github.com/emiliantolo/pytorch\textunderscore nsfw\textunderscore model}} model uses the backbone fine-tuned on around 40GB of \emph{nsfw} data .
We divide our model for multi-label classification according to different types of loss used during the training stage.

\begin{table}[]
  \centering
  \resizebox{\linewidth}{!}{
    \begin{tabular}{lc|c|c}\toprule[2pt]
                              &                    & \multicolumn{1}{c|}{\textbf{Binary}} & \multicolumn{1}{c}{\textbf{Multi-Label}} \\\cmidrule(lr){3-3}\cmidrule(lr){4-4}
      \textbf{Backbone Model} & \textbf{$\lambda$} & $f_{1}^{macro}$                      & $f_{1}^{weighted}$                       \\
      \midrule[2pt]
      ResNet-10$1_{imagenet}$   & 0.1                & 0.601                               & 0.590                                      \\
                              & 0.2                & 0.619                               & 0.597                                      \\
                              & 0.3                & 0.645                               & 0.622                                     \\
                              & 0.4                & 0.689                               & 0.628                                      \\
                              & 0.5                & 0.702                               & 0.641                                      \\
                              & 0.6                & 0.736                               & \textbf{0.645}                                      \\
                              & 0.7                & \textbf{0.741}                              & 0.643                                      \\
                              & 0.8                & 0.728                               & 0.631                                      \\
                              & 0.9                & 0.702                               & 0.612                                      \\
      \hline
      ResNet-50$_{nsfw}$        & 0.1                & 0.611                               & 0.591                                      \\
                              & 0.2                & 0.620                               & 0.595                                      \\
                              & 0.3                & 0.691                               & 0.612                                     \\
                              & 0.4                & 0.703                               & 0.632                                      \\
                              & 0.5                & 0.736                               & 0.634                                      \\
                              & 0.6                & 0.749                               & \textbf{0.635}                                      \\
                              & 0.7                & \textbf{0.759}                              & 0.632                                      \\
                              & 0.8                & 0.734                               & 0.623                                      \\
                              & 0.9                & 0.698                               & 0.601                                      \\
      \bottomrule
    \end{tabular}
  }
  \caption{$\lambda$ effect on model performance }
  \label{tab:lambda}
\end{table}

\begin{table}[h]
    \centering
    \begin{tabular}{lc}\toprule[2pt]
     \multicolumn{2}{c}{\textbf{SubTask-A}} \\\midrule
     Model & $f_{1}^{macro}$ \\\midrule[1pt]
     Baseline-Text & 0.640 \\
     Baseline-Image & 0.639 \\
     Baseline-IT & 0.543 \\
     \text{Ours(ResNet-101$_{imagenet}$)} & 0.751\\
     \text{Ours(ResNet-50$_{nsfw}$)} & \textbf{0.759} \\ \bottomrule
\end{tabular}
    \caption{Comparing the  $f_{1}^{macro}$ of our methods and the baselines for binary classification task. }
    \label{tab:binresults}
\end{table}

\begin{table}[]
    \centering
    \begin{tabular}{lr|c}\toprule[2pt]
     \multicolumn{3}{c}{\textbf{SubTask-B}} \\\midrule
     \multicolumn{2}{c|}{Model} & $f_{1}^{weighted}$ \\\midrule[1pt]
     \multicolumn{2}{c|}{Baseline-Flat} & 0.421 \\
     \multicolumn{2}{c|}{Baseline-Heirarchical} & 0.621 \\
     \multicolumn{2}{c|}{\text{Ours(ResNet-101$_{imagenet}$)}} & \\ 
     & \small{$+$ \texttt{SM Loss}} & 0.641 \\
     & \small{$+$ \texttt{Custom Loss}} & \textbf{0.645} \\
     \multicolumn{2}{c|}{\text{Ours(ResNet-50$_{nsfw}$)}} & \\ 
     & \small{$+$ \texttt{SM Loss}} & 0.632 \\
     & \small{$+$ \texttt{Custom Loss}} & 0.638 \\\bottomrule
    \end{tabular}
    \caption{Comparing the  $f_{1}^{weighted}$ of our methods and the baselines for multi-label classification task.\texttt{SM Loss} refers to multi-label SoftMargin loss}
    \label{tab:multiresults}
\end{table}
\subsection{Task Results}

\textbf{Subtask-A}: The results of the experiments for the binary classification task can be seen in Table \ref{tab:binresults}. For Subtask-A, the models that used
multi-model training and and an additional navigator network on the image end outperformed the single modality models and the simple multimodal concatenation model.
Amongst the model using \emph{ResNet} backbone , the model fine-tuned on \emph{nsfw} had an edge over the model which had been pre-trained on \emph{imagenet} dataset. The can indicate that there is an indicator of women's image representation with the meme being a misogynstic one .

\textbf{Subtask-B}: The results of the experiments for the multi-label classification task can be seen in Table \ref{tab:multiresults}. For Subtask-B, the models that used graph network to create independent classifiers and an additional navigator network on the image end outperformed the models using the simple multimodal concatenation model with classification head and the heirarchical multi-label model using text representations . Amongst the model using \emph{ResNet} backbone , the model fine-tuned on \emph{nsfw} dataset performed poorer to the model which had been pretrained on \emph{imagenet} dataset. This can be an indicator that general feature representations are perhaps more important for identification of the specificity of misogyny as compared to that of the fine-tuned feature feature representations.

\subsection{Ablation Studies}
In this section, we perform ablation studies from two different aspects,particularly including the sensitivity of the classification models to effects of $\lambda$ when concatenating the two different types of modalities , visual and textual together , to determine the relative importance of the two with respect to each other , and the other being the depths of Graph Classification Module which we use for the multi-label classification model.

\textbf{Effects of different threshold values $\lambda$ :} We vary the values of the threshold concatenation parameter $\lambda$ from 0.1 to 0.9 in steps of 0.1 . $\lambda = 0$ corresponds to building the entire feature vector from the visual stream while $\lambda = 1$ corresponds to entire information coming from the textual stream. The results are shown in table \ref{tab:lambda} , where the performance of the two models based on ResNet-101 backbone are compared pretrained on two different datasets. It can be observed that the textual stream information is of higher importance in both the classification problem as the performance boost is skewed for roughly $\lambda  = [0.6 , 0.7]$. It may be due to the fact that in the images as well , a good amount of information that is used to recognise the misogyny of the meme is cognitively of the textual nature , while the image content of the meme is lesser in comparison to its textual counterpart. It may also be that the image content is not of high quality.

\begin{table}[]
    \centering
    \resizebox{\linewidth}{!}{
    \begin{tabular}{lcc}\toprule[2pt]
    & & \multicolumn{1}{c}{\textbf{Multi-Label}}\\\cmidrule(lr){3-3}
               \textbf{Backbone Model} & \textbf{Graph Depth}  & $f_{1}^{weighted}$ \\
    \midrule[2pt]
    ResNet-101$_{imagenet}$    & \textsl{2-layer} & \textbf{0.644}   \\
     & \textsl{3-layer} & 0.644   \\
     & \textsl{4-layer} & 0.632  \\
     & \textsl{5-layer} & 0.628   \\
     \hline
    ResNet-50$_{nsfw}$     & \textsl{2-layer} & 0.641   \\
     & \textsl{3-layer} & \textbf{0.643}   \\
     & \textsl{4-layer} & 0.629  \\
     & \textsl{5-layer} & 0.621   \\
    \bottomrule
    \end{tabular}
}
    \caption{Graph Network Depth effect on model performance }
    \label{tab:gcn}
\end{table}

\textbf{Effects of different depth of Graph Classification Network:} We vary the values of the the number of layers of the graph network from 2 to 5 and observe its effect on the model performance. For the two layer model , the output dimensions of the layers are 512, 2048 , for the three-layer model, the output dimensionalities are 512, 1024 and 2048 for the sequential layers , for the four-layer model, the dimensionalities are 512, 1024, 1024 and 2048 , and for the five-layer model , the output dimensions are 512, 1024, 1024, 1024, 2048.As shown in table \ref{tab:gcn}, when the number of graph convolution layers increases, multi-label recognition performance drops on both datasets. The possible reason for the performance drop may be that when using more GCN layers, the propagation between nodes will be accumulated, which can result in over-smoothing.

\textbf{Effects of using Custom Loss Function:} We compare the results for multi-label classification with two types of losses : (i) MultiLabel Soft Margin Loss (\texttt{SM Loss}); (ii) Custom Loss as described in \ref{subsec:customloss}. From table \ref{tab:multiresults} , we can see that the \texttt{Custom Loss} out performs the \texttt{SM Loss} in the experimental runs .

The result can be explained to the fact that weighted classes that affect the loss value for positive as well as negative labels. 

\textbf{(i)} If the model predicts a positivity for the label which has higher positive weightage the loss value would increase , thereby forcing the model to not favour one particular label. Similarly when the model predicts negative value for a particular label which has higher negative weightage , the loss value would increase , forcing the model to not favour negativity of a particular label.

\textbf{(ii)} If the model predicts a positivity for the label which has lower positive weightage, the loss value would decrease , thereby forcing the model predict favourably for that particular label. Similarly when the model predicts negative value for a particular label which has lower negative weightage , the loss value would decrease , forcing the model to favour negativity of that particular label.

\section{Conclusion}
We have described the systems developed by as to solve the Multimedia Automatic Misogyny Identification challenge at Semeval 2022 . In our best performing submission for SubTask-A , we framed the problem as a binary classification task and used two seperate streams of information simulataneously to identify misogyny , while for our model for SubTask-B , we tried to find the semantic relation between the type of misogyny and their relative importance to solve the problem for Multi-Label classification. By making use of powerful, state-of-the-art, pretrained models for text and images, our models were able to achieve high F1 score for both the tasks . Our best performing model ranked 7th out of 83 participants submissions on SubTask-A and 38th out of 83 participants on SubTask-B.

As part of future work, we aim to explore alternate approaches to model the multi-label dependencies using Knowledge-Graph and GAT Networks. Also , there seems to be a problem of oversmoothing when increasing the depth of the Graph Classification Module , which we aim to resolve using effective Normalization layers between the graph layers.

\bibliography{anthology,custom}
\bibliographystyle{acl_natbib}

\appendix

\end{document}